\documentclass[cameraready]{Interspeech}

\usepackage{siunitx}
\usepackage{makecell}
\usepackage{multirow}
\usepackage{tabularx}
\usepackage{adjustbox}

\newcommand{\apptek}{%
  \ifcameraready
    AppTek%
  \else
    ANONYMIZED-ORG-NAME%
  \fi
}
\newcommand{\appteksmall}{%
  \ifcameraready
    AppTek%
  \else
    REDACTED%
  \fi
}
\newcommand{\AT}{%
  \ifcameraready
    AT%
  \else
    RD%
  \fi
}

\newcolumntype{n}{>{\sisetup{
  detect-all,
  mode=text,
  table-format=3.1,
  parse-numbers = true,
  input-ignore = {},
  table-align-text-post = false
}}S}
\newcommand{\rotheader}[1]{%
  \multicolumn{1}{c}{%
    \adjustbox{angle=40,lap=\width-(\tabcolsep)-5mm}{\textbf{#1}}%
  }%
}
\newcommand{\best}[1]{\bfseries #1 \mdseries}
\title{\apptek\ Call-Center Dialogues: A Multi-Accent Long-Form Benchmark for English ASR\thanks{Submitted to INTERSPEECH 2026.}}

\author[orcid=0000-0003-1641-6628, equalcontribution, correspondingauthor]{Eugen}{Beck}
\author[orcid=0009-0007-3565-5620, equalcontribution, correspondingauthor]{Sarah}{Beranek}
\author[orcid=0009-0008-5043-2393, equalcontribution, correspondingauthor]{Uma}{Moothiringote}
\author[orcid=0009-0001-6239-3650]{Daniel}{Mann}
\author[orcid=0009-0004-6962-7562]{Wilfried}{Michel}
\author[orcid=0009-0008-1749-3928]{Katie}{Nguyen}
\author[orcid=0009-0002-3802-0421]{Taylor}{Tragemann}

\address{
    AppTek.ai
}

\email{\{ebeck,sberanek,umoothiringote\}@apptek.com }

\keywords{automatic speech recognition, accent robustness, long-form speech, conversational speech, call-center dialogue, speech dataset, ASR evaluation}

\begin{document}

\maketitle

\setcounter{footnote}{0}
\pagenumbering{arabic} 
\renewcommand{\thefootnote}{\arabic{footnote}} 

\begin{abstract}
Evaluating English ASR systems for conversational AI applications remains difficult, as many publicly available corpora are either pre-segmented into short segments, consist of read or prepared speech, or lack explicit dialect annotations to evaluate robustness for a diverse user base. This work presents the \apptek\ Call-Center Dialogues corpus, a collection of spontaneous, role-played agent-customer conversations spanning fourteen English accents covering sixteen service-oriented scenarios. The dataset was commissioned specifically for evaluation and none of the audio or text was publicly available prior to release, reducing the risk of overlap with existing large-scale pretraining corpora. We benchmark a set of open-source ASR systems under different segmentation approaches. Results show substantial variation across accents and segmentation methods, indicating that good performance on general American English benchmarks does not necessarily generalize to other accents.
\end{abstract}

\vspace{-0.2cm}
\section{Introduction}
\label{sec:introduction}

Recent advances in automatic speech recognition (ASR) have led to strong performance on standard English benchmarks. However, systematic evaluation under realistic long-form conversational conditions remains limited, particularly across diverse accents, as most public benchmarks emphasize pre-segmented recordings of read or prepared speech rather than spontaneous and interactive speech. This gap is especially relevant for use cases in the area of conversational AI, e.g. automated call centers, where systems must process extended interactions containing disfluencies, repairs, named entities, and domain-specific vocabulary.

A second challenge is benchmark integrity for large open-weight ASR models that often do not fully disclose which training data went into their creation. If large-scale web scraping was performed to gather the data, the transcripts of publicly available test sets, or near-duplicates thereof could have ended up in the training. Two pragmatic mitigations are (i) conducting private evaluations with data that is not publicly accessible, and/or (ii) periodically refreshing test material so the benchmark remains out-of-distribution as training pipelines evolve.

With this work we chose the latter approach and present a corpus of spontaneous, role-played call-center dialogues spanning fourteen English accents, covering sixteen topics. It was created specifically for evaluation purposes. Verbatim transcripts were produced by professional annotators following a protocol with multiple stages of quality assurance. We also provide word error rates by accent and in aggregate under multiple segmentation strategies for multiple open-weight ASR systems. The corpus is available as a Hugging Face dataset\footnote{\scriptsize\url{https://huggingface.co/datasets/apptek-com/apptek_callcenter_dialogues}} under Creative Commons 4.0 BY-SA \footnote{\scriptsize\url{https://creativecommons.org/licenses/by-sa/4.0/}}.
To the best of our knowledge, this corpus represents the largest publicly available collection of English-accented conversational speech recorded and transcribed under comparable conditions.

\vspace{-0.2cm}
\section{Related Work}
\label{sec:related_work}

Progress in ASR has been strongly shaped by evaluation on widely used benchmarks, many of which contain short, read speech, such as LibriSpeech~\cite{panayotov2015librispeech}, Mozilla Common Voice~\cite{ardila2020common}, and FLEURS~\cite{conneau2022fleurs}. These datasets reduce transcription effort because the text intended to be spoken is known, but they are less representative of deployed conversational ASR and can bias evaluation toward systems with strong language modeling, especially those leveraging pretrained LLMs that may have seen the underlying text during training. They also under-represent phenomena common in spontaneous speech, such as false starts, disfluencies, repetitions or hesitations.

To better capture spontaneous speech, several corpora focus on naturally produced conversational audio, including Hub5~\cite{ldc2002t43}, AMI~\cite{carletta2006}, and CHiME-6~\cite{barker18}. However, these datasets are designed for different target conditions, namely telephone conversation, meeting transcription, and far-field speech in noisy homes, respectively, and therefore emphasize different challenges such as channel effects, multi-party overlap, and noise or reverberation. While valuable, they do not directly reflect the long-form, task-oriented agent-customer interactions characteristic of call-center applications, nor do they provide explicit global English accent coverage for systematic accent-robustness evaluation in this setting.

The dataset closest to our effort is Earnings-22~\cite{earnings22}, which contains earnings calls from global companies and thus includes a range of English accents. Because accent is difficult to label externally and calls involve multiple speakers, the authors group the data by the company’s associated country; this weakens conclusions about specific accents since speakers may not match the company location and may themselves be diverse. Moreover, earnings calls include substantial read or prepared speech, especially initial statements, and typically limited interaction, differing from the extended, interactive call-center dialogues considered here. Finally, because the audio and transcripts are sourced from public material, large foundational models may have been trained on the data. 

\begin{table*}[ht]
\caption{Dataset statistics and speaker demographics by accent, including number of speakers, recordings, and speech duration.}
\label{tab:dataset_by_accent}
\centering
\begin{tabularx}{17cm}{ l l r r r r r r r r }
\toprule
  \multicolumn{2}{c}{\textbf{Accent}} & \multicolumn{3}{c}{\textbf{Age}} 
& \multicolumn{2}{c}{\textbf{Gender}} 
& {\multirow{2}{*}{\textbf{\#Speakers}}} & {\multirow{2}{*}{\textbf{\#Calls}}} & {\multirow{2}{*}{\textbf{Duration [h]}}} \\
\cmidrule(lr){1-2} \cmidrule(lr){3-5} \cmidrule(lr){6-7} 
\textbf{Code} & \textbf{Name} & \textbf{18--30} & \textbf{30--50} & \textbf{50--70} 
& \textbf{Female} & \textbf{Male} 
& & & \\
\midrule
en\_AU          & Australian                                   & 4  & 5  & 1  & 7  & 3  & 10 & 58 &  9.1 \\
en\_CA          & Canadian                                     & 4  & 5  & 1  & 7  & 3  & 10 & 59 &  8.8 \\
en\_CN          & Chinese                                      & 11 & 0  & 0  & 6  & 5  & 11 & 79 &  8.4 \\
en\_GB          & British                                      & 5  & 3  & 2  & 9  & 1  & 10 & 67 & 10.7 \\
en\_GB\_SCT     & Scottish                                     & 6  & 2  & 4  & 9  & 3  & 12 & 66 &  9.1 \\
en\_GB\_WLS     & Welsh                                        & 5  & 3  & 2  & 6  & 4  & 10 & 65 &  9.5 \\
en\_IE          & Irish                                        & 6  & 1  & 3  & 4  & 6  & 10 & 56 &  9.6 \\
en\_IN          & Indian                                       & 10 & 2  & 1  & 6  & 7  & 13 & 73 & 10.0 \\
en\_MX          & Mexican                                      & 2  & 6  & 2  & 4  & 6  & 10 & 61 &  8.8 \\
en\_SG          & Singaporean                                  & 4  & 6  & 0  & 5  & 5  & 10 & 57 &  8.1 \\
en\_US\_AAVE    & \makecell[l]{African American \\ Vernacular} & 4  & 4  & 2  & 9  & 1  & 10 & 45 &  8.5 \\
en\_US\_General & General US American                          & 6  & 6  & 2  & 10 & 4  & 14 & 67 &  9.3 \\
en\_US\_South   & Southern US American                         & 0  & 6  & 4  & 7  & 3  & 10 & 56 &  9.2 \\
en\_ZA          & South African                                & 9  & 7  & 0  & 13 & 3  & 16 & 64 &  9.4 \\
\midrule
\textbf{Total}  &  & 76 & 56 & 24 & 102 & 54 & 156 & 873 & 128.6 \\
\bottomrule
\end{tabularx}
\end{table*}

\vspace{-0.2cm}
\section{Dataset description}
\label{sec:dataset}

\subsection{Overview}
The dataset is an English ASR test set of spontaneous role-played call-center conversations spanning fourteen English accents and multiple service-oriented domains. It is designed exclusively for evaluation and analysis rather than model training. In total, the corpus contains 128.6 hours of speech across 156 speakers and 1,746 single-channel recordings, with approximately 8-11 speech hours per accent. The speech duration is derived from annotated transcription segments. Detailed statistics are provided in Table~\ref{tab:dataset_by_accent}.

\vspace{-0.2cm}
\subsection{Transcription Conventions}
\label{sec:transcription_convention}

Speech was transcribed verbatim, preserving disfluencies, repetitions, and conversational repairs. Filled pauses were retained and marked with a \# symbol (\texttt{\#um}); false starts and truncated words were marked with a tilde (\texttt{he$\sim$ hello}). Non-standard grammatical usage was preserved. Dialectal pronunciations were rendered using standard orthography rather than phonetic spelling, while dialect-specific lexical items were retained when part of natural usage (e.g., discourse particles in Singapore English).
Numerals and symbols were written in spoken form (\texttt{five dollars}). Acronyms and abbreviations were transcribed according to pronunciation (\texttt{NASA}, \texttt{professor}), with limited exceptions for \texttt{Ms, Mr, Mrs, Mx}. Initialisms were marked using \texttt{<initial>...</initial>}. Non-English spans were tagged as \texttt{<lang:X>...</lang:X>}, and unintelligible regions were marked using \texttt{(())}. Spelling, punctuation, and casing followed U.S. English conventions, even for accents that regularly use British English conventions.

\vspace{-0.2cm}
\subsection{Speaker Recruitment and Demographics}
 \label{sec:speaker_selection}
Speakers were recruited through established call-center partners, freelance contributors, and voice data collaborators with prior experience in speech collection. Fourteen accent categories were defined to represent commonly recognized global varieties of English. Participants were required to be at least 18 years old and native to the target locale (minimum second generation in-region).
Accents were self-identified and verified through sample recordings and a structured onboarding process. Natural intra-accent variation was accepted, and while accents are treated as discrete evaluation categories, strict linguistic boundaries were not imposed. The dataset includes 10–16 speakers per accent, with no overlap across accent groups.

Recruitment aimed to encourage demographic diversity. Speakers span a broad age range (49\% aged 18–30, 36\% aged 30–50, and 15\% aged 50–70), with a gender distribution of 65\% female and 35\% male, see Table~\ref{tab:dataset_by_accent}.

\vspace{-0.2cm}
\subsection{Recording Setup and Conditions} 
\label{sec:recording_setup_and_conditions}
Conversations were collected as paired, role-based agent–customer dialogues in a free-form, spontaneous manner without scripted text. Limited pre-session planning, where speakers selected a service-oriented topic and aligned on a general scenario, was permitted, but participants were instructed to use their natural speaking style. Disfluencies and restarts were expected.
Dialogues were recorded in sessions ranging from 5 to 15 minutes (10.4 minutes on average). Speakers could participate in multiple sessions, contributing up to approximately one hour of transcribed speech.

Topics span a wide range of domains: agriculture, aviation, banking, delivery service, energy, entertainment, finance, food, health, hospitality, insurance, real estate, retail, technology, telecommunication, and travel. Dialogues frequently contain named entities and numerical expressions (e.g. dates, account numbers, billing amounts), reflecting realistic call-center interactions and supporting evaluation under domain-specific vocabulary variation. To protect privacy, participants were instructed to use fictional but plausible entities when needed. Inappropriate content and excessive profanity were discouraged.

Recordings were conducted via a VoIP-based platform and exported as 16 kHz, 16-bit linear PCM WAV split-channel audio files (one channel per speaker). Sessions were recorded using consumer devices, primarily laptops (53\%), phones (42\%), and tablets (5\%). Automated quality checks, including clipping detection, level monitoring, and post-recording SNR verification, were applied. Most recordings were completed in quiet home environments (78\%), with some in controlled indoor public spaces (19\%) and rarely outdoors (3\%). Light background noise was permitted if speech remained clearly intelligible.

\subsection{Transcription Process}
\label{sec:transcription}

All recordings were manually transcribed by professional annotators without help of any automatic tools to bootstrap or pre-generate transcripts. The corpus follows a verbatim transcription protocol designed to preserve conversational phenomena common in spontaneous dialogue, see section~\ref{sec:transcription_convention}. In total, 85 annotators participated. Annotators were required to be native English speakers or demonstrably familiar with the assigned accent group.

Transcripts were produced in a first-pass transcription stage followed by multi-round quality assurance (QA). After initial transcription and segmentation by an annotator, files underwent review by experienced QA-annotators. A subset of QA-reviewed files was further evaluated by senior annotators or project managers for final approval. Automated validation checks detected formatting errors, invalid characters, and segmentation inconsistencies. 

In addition to manual quality assurance, a targeted automatic consistency check was applied to flag segments for manual re-review. This procedure employed \textit{guided recognition}, in which a 4-gram background language model (LM) was linearly interpolated with a segment-specific LM estimated from the proposed transcription. The resulting LM biases recognition toward the transcript while still permitting deviations supported by strong acoustic evidence or high linguistic deviation.

The guided-recognition hypothesis was compared to the transcription using Levenshtein alignment. A segment was flagged for manual review if (i) the word-level edit distance between hypothesis and transcription was larger than a threshold ($n\ge 4$), and (ii) the recognition system was confident in its deviation assessment with the minimum confidence over all words in the sequence higher than a threshold ($c\ge 0.56$).

The thresholds were chosen such that 10\% of all segments were flagged. These segments were returned for manual re-evaluation without providing the deviating hypothesis. In this sample, approximately 40\% of flagged segments contained minor transcription issues, which were fixed in the additional QA round.

\subsection{Multilingual Extension}

A subset of the dataset (approximately five hours of speech) was professionally translated into Chinese, German, Japanese and Spanish. These translations will serve as blind evaluation data 
\ifcameraready
at the 2026 shared task of the International Workshop on Spoken Language Translation (IWSLT).
\else
at an upcoming machine translation workshop.
\fi
The translated subset is derived from the same conversational recordings, including speakers from the US, Canada, India, China, and other accents, and follows the original segmentation.

\vspace{-0.2cm}
\section{Benchmarking}

\subsection{Evaluation Setup}

A diverse set of publicly available open-weight ASR systems were evaluated on the test set. All models were executed locally using their  default inference settings. The evaluated models are 
NVIDIA Canary-1B v2, Parakeet 0.6B TDT (v2, v3)~\cite{sekoyan2025canary1bv2parakeettdt06bv3efficient} and NeMo Canary-Qwen-2.5B~\cite{canary_qwen},
IBM Granite Speech 3.3 (2B and 8B)~\cite{saon2025granitespeechopensourcespeechawarellms}, 
Kyutai STT 2.6B en~\cite{kyutai2025streaming},
Microsoft Phi-4 Multimodal Instruct~\cite{microsoft2025phi4minitechnicalreportcompact}, 
Alibaba Qwen3-ASR (0.6B and 1.7B)~\cite{Qwen3-ASR} and
OpenAI Whisper Large (v2, v3)~\cite{whisper}.

Since most evaluated models require short utterances to work, recognition was performed using different segmentation strategies: Manual segmentation~(Man.), \apptek{}s proprietary segmenter~(\AT), Silero segmenter~(Sil.)~\cite{Silero_VAD},\footnote{Silero Settings: $\mbox{min}\_\mbox{silence}\_\mbox{duration}=10.0$,\\  $\mbox{min}\_\mbox{speech}\_\mbox{duration}=0.25$ and $\mbox{max}\_\mbox{speech}\_\mbox{duration}=30$} and fixed-length chunking with 30s and 60s windows. For reference, the average segment lengths and standard deviations are shown in Table~\ref{tab:baseline_models}. All models were evaluated with identical segmentation.

Recognition performance was measured using word error rate (WER). Although recognition was done on segmented audio, scoring was aggregated per-session to reflect full conversational interactions. Scoring follows the Hugging Face Open\-ASR leaderboard protocol~\cite{srivastav2025openasrleaderboardreproducible}, including case normalization, punctuation removal, and number normalization. To ensure consistent scoring across models with differing output formats, a dataset-specific normalization was additionally applied prior to evaluation, which reduced WER by approximately 0.8-1.1\% absolute consistently across all models and test sets. The normalization script is part of the dataset publication. 

\begin{table}[t]
  \caption{WER (\%) across different segmentation strategies, averaged over all 14 accents, sorted by model size. Whisper cuts input audio after 30s. \\Man = Manual, \AT = \appteksmall, Sil = Silero, Fixed = fixed-length chunking}
  \label{tab:baseline_models}
  \centering
  \begingroup
  \setlength{\tabcolsep}{4pt}
  \begin{tabular}{ l  r  r  r  r  r }
    \toprule
    \textbf{Model} & \textbf{Man.} & \textbf{\AT} & \textbf{Sil.} & \multicolumn{2}{c}{\textbf{Fixed}}\\ 
    \midrule
    avg. segment len.
        & $4.9$s & $7.9$s & $16.5$s & 30.0s & 60.0s \\
    $\pm$ std.
    & $\pm3.7$s & $\pm8.7$s & $\pm9.6$s & - & - \\
    \midrule
    Parakeet v2 
        & \textbf{9.2} & 9.5 & 9.6 & 10.1 & 9.4 \\
    Parakeet v3 
        & \textbf{8.8} & 9.0 & 9.2 & 9.9 & 12.1 \\
    Qwen3-ASR 0.6B
        & 8.9 & 8.9 & 9.2 & 8.9 & \textbf{8.7} \\
    Canary-1B v2 
        & \textbf{10.6} & 11.2 & 11.2 & 10.9 & 13.3 \\
    Whisper Large v2 
        & 18.5 & 26.9 & \textbf{16.0} & 48.4 & -- \\
    Whisper Large v3 
        & \textbf{10.7} & 18.9 & 15.0 & 42.9 & -- \\
    Qwen3-ASR 1.7B
        & 7.9 & 8.0 & 8.3 & 7.8 & \textbf{7.4} \\
    Granite 2B 
        & \textbf{10.8} & 11.6 & 13.1 & 14.0 & 19.7 \\
    Canary-Qwen 2.5B
        & \textbf{8.6} & 9.2 & 9.2 & 8.9 & 10.0 \\
    Kyutai STT 2.6B
        & \textbf{11.1} & \textbf{11.1} & 11.3 & 12.1 & 13.2 \\
    Phi-4 Multimodal 
        & \textbf{9.2} & 9.8 & 10.0 & 11.9 & 18.8 \\
    Granite 8B 
        & \textbf{10.5} & 10.9 & 11.9 & 12.2 & 13.8 \\
    \bottomrule
  \end{tabular}
  \endgroup
\end{table}

\begin{table*}[t]
\caption{WER (\%) by English accent across evaluated models sorted by model size using the silero segmenter setup.}
\label{tab:accent_models_transposed}
\centering  
\begingroup
\setlength{\tabcolsep}{5.5pt}
\begin{tabularx}{16.7cm}{l  n n n n n n n n n n n n | n}
\toprule
\textbf{Accent}
& \rotheader{\textbf{Parakeet v2}}
& \rotheader{\textbf{Parakeet v3}}
& \rotheader{\textbf{Qwen3-ASR}}
& \rotheader{\textbf{Canary-1B}}
& \rotheader{\textbf{Whisper v2}}
& \rotheader{\textbf{Whisper v3}}
& \rotheader{\textbf{Qwen3-ASR}}
& \rotheader{\textbf{Granite}}
& \rotheader{\textbf{Canary-Qwen}}
& \rotheader{\textbf{Kyutai STT}}
& \rotheader{\textbf{Phi-4}}
& \rotheader{\textbf{Granite}} 
& \rotheader{\textbf{Avg.}}\\
\midrule
Model size
& {0.6B} & {0.6B} & {0.6B}
& {1B}
& {1.6B} & {1.6B}
& {1.7B}
& {2B}
& {2.5B}
& {2.6B}
& {5.6B}
& \multicolumn{1}{c}{{8B}} 
& \\
\midrule
en\_AU          & 5.6 & 5.2 & 5.3 & 6.6 & 9.3 & 8.1 & \best{4.7} & 6.4 & 5.2 & 5.9 & 5.4        & 6.2 & 6.2\\
en\_CA          & 8.3 & 7.6 & 7.3 &10.1 &16.4 &14.5 & \best{6.9} &12.6 & 8.2 & 8.6 & 8.1        &10.3 & 9.9\\
en\_CN          &12.6 &12.9 &11.7 &14.7 &18.2 &20.1 &\best{10.3} &18.8 &12.1 &15.6 &13.3        &14.8 &14.6\\
en\_GB          &10.4 &10.3 &10.0 &11.5 &14.2 &16.6 & \best{9.2} &13.5 &10.1 &11.3 &11.0        &12.2 &11.7\\
en\_GB\_SCT     &12.4 &12.1 &12.3 &14.3 &17.4 &17.3 &\best{11.1} &16.5 &12.3 &14.1 &13.2        &15.8 &14.1\\
en\_GB\_WLS     &10.7 &10.7 &10.3 &12.2 &16.6 &16.6 & \best{9.5} &13.2 &10.4 &11.8 &11.2        &12.1 &12.1\\
en\_IE          & 8.1 & 7.3 & 7.6 & 9.6 &12.8 &13.0 & \best{6.6} &11.4 & 8.3 & 9.5 & 8.8        &10.0 & 9.4\\
en\_IN          & 9.9 & 9.7 &11.0 &12.9 &33.0 &11.9 &       10.3 &18.8 & 9.5 &14.9 & \best{9.4} &15.7 &13.9\\
en\_MX          &10.9 &10.9 &10.3 &12.2 &14.3 &18.4 & \best{9.3} &13.2 &10.6 &13.2 &10.9        &12.6 &12.2\\
en\_SG          &12.4 &12.4 &12.4 &14.9 &15.9 &18.0 &\best{10.9} &19.1 &12.1 &16.5 &14.3        &18.8 &14.8\\
en\_US\_AAVE    & 9.0 & 8.1 & 8.2 & 9.9 &14.6 &15.3 & \best{7.2} &11.4 & 7.9 & 9.4 & 9.2        &10.7 &10.1\\
en\_US\_General & 6.2 & 5.5 & 5.6 & 7.6 &11.0 & 9.9 & \best{5.0} & 7.9 & 5.8 & 6.7 & 6.2        & 7.5 & 7.1\\
en\_US\_South   & 7.8 & 7.1 & 7.2 & 8.7 &13.7 &12.1 & \best{6.4} &10.4 & 7.0 & 8.4 & 7.7        & 9.1 & 8.8\\
en\_ZA          &10.1 & 9.6 & 9.8 &11.4 &16.2 &19.1 & \best{8.9} &12.7 & 9.8 &12.5 &10.8        &11.4 &11.9\\
\midrule
Avg.            & 9.6 & 9.2 & 9.2 &11.2 &16.0 &15.0 & \best{8.3} &13.3 & 9.2 &11.3 &10.0 &\multicolumn{1}{n}{11.9} & \\
\bottomrule
\end{tabularx}
\endgroup
\end{table*}


\subsection{Results}

Table~\ref{tab:baseline_models} reports WER averaged across all fourteen accents under each segmentation strategy. Manual segmentation yields the best performance for nearly all evaluated systems, indicating that accurate boundary detection remains crucial for long-form conversational ASR. The primary exception are the Qwen3-ASR models, which achieve their lowest WER under fixed 60 s chunking, suggesting greater robustness to longer unstructured inputs. Inference without external segmentation resulted in meaningful results only for Kyutai STT 2.6B (13.9\% WER) and NVIDIA Parakeet 0.6B TDT (v2: 8.8\%, v3: 10.4\%).

To analyze accent robustness, Table~\ref{tab:accent_models_transposed} reports WER by accent using the Silero VAD setup. Substantial variation persists across accents. For several models, the WER gap between the lowest- and highest-performing accents exceeds 10\% absolute. Accents such as en\_SG, en\_CN, en\_GB\_SCT and en\_IN consistently yield higher error rates across systems, whereas en\_AU and en\_US\_General tend to achieve lower WER. 
We notice that the relative gap between the best and the worst performing accent does not correlate with average model performance, e.g. the relative difference for Canary-1B is 26\% at an average WER of 11.2\% whereas for Parakeet V3 the relative gap is 48\% at an average WER of 9.2\%.
This suggests that improvements in average recognition accuracy do not automatically translate into accents robustness.

Overall, the results demonstrate that long-form conversational speech and accent diversity jointly introduce challenges that are not fully captured by short-form or read speech benchmarks~\cite{srivastav2025openasrleaderboardreproducible}. The observed sensitivity to segmentation further underscores the importance of evaluation protocols aligned with realistic deployment conditions.

\vspace{-0.2cm}
\section{Limitations}

The dataset is restricted to role-played call-center interactions. Thus participants might not be familiar with all technical terms or expressions used in a given domain.

While demographic diversity was encouraged during recruitment, gender distribution is not balanced across all accent groups~(see Table~\ref{tab:dataset_by_accent}). In total, 102 female and 54 male speakers participated, with certain accents exhibiting stronger imbalance (e.g. en\_GB, en\_US\_AAVE, en\_ZA), while others are more balanced (e.g. en\_IN, en\_SG). Such imbalance may influence acoustic variability and should be taken into account.

Accent labels are self-reported and subsequently verified by in-house reviewers. However, accent categories are treated as discrete evaluation groups despite natural intra-accent variation. Certain regions exhibit internal dialectal diversity that is not fully represented. For example, South African English encompasses multiple regional influences; in this dataset, most South African speakers report Zulu as their primary native language, with limited Afrikaans representation, while Canadian English primarily reflects speakers from English-dominant regions. Results should therefore be interpreted as performance on the represented speaker sample rather than exhaustive coverage of each accent community.

Verbatim transcription of spontaneous, accented conversational speech is inherently challenging, particularly for rapid speech and reduced articulation. Although multi-stage quality assurance was applied, no formal inter-annotator agreement metric was computed. Residual transcription uncertainty may therefore remain, especially in acoustically challenging segments.

\vspace{-0.2cm}
\section{Conclusion}

This work introduced the \apptek\ Call-Center Dialogues, a long-form English ASR test set of spontaneous, role-played agent-customer conversations spanning fourteen English accents and sixteen service-oriented scenarios. The dataset was collected from scratch and does not rely on publicly available sources, minimizing potential overlap with web-scraped training data. The dataset contains 129 hours of transcribed speech, which represents, to the best of our knowledge, the largest publicly available collection of English-accented conversational speech recorded and transcribed in a controlled and comparable setting. Together with the released evaluation protocol, the corpus enables reproducible benchmarking under realistic conversational conditions and supports systematic analysis of ASR performance across accents, gender, and other demographic factors relevant to conversational AI deployments.

Benchmarking across a range of recent open-weight ASR models revealed substantial sensitivity to both accent and segmentation strategy. Manual segmentation consistently yielded the lowest WER for most systems, indicating that robust boundary detection remains a critical component for long-form conversational ASR. Across accents, error rates varied widely and the gap between best- and worst-performing accents remained large even for strong average-performing models, suggesting that improvements in overall WER do not automatically translate into accent robustness.

\clearpage

\section{Generative AI Use Disclosure}

OpenAI's ChatGPT (GPT5.2~\cite{singh2025openaigpt5card}) was used to proofread the paper.
The gpt-oss-120B~\cite{openai2025gptoss120bgptoss20bmodel} model was used locally to help generate the mapping files for scoring normalization and to verify proper US English spelling.
Any generative AI output was vetted by at least one of the authors before including it in this work.

\bibliographystyle{IEEEtran}
\bibliography{mybib}

@misc{Qwen3-ASR,
    title={{Qwen3-ASR Technical Report}}, 
    author={Xian Shi and Xiong Wang and Zhifang Guo and Yongqi Wang and Pei Zhang and Xinyu Zhang and Zishan Guo and Hongkun Hao and Yu Xi and Baosong Yang and Jin Xu and Jingren Zhou and Junyang Lin},
    year={2026},
    eprint={2601.21337},
    archivePrefix={arXiv},
    primaryClass={cs.CL},
    url={https://arxiv.org/abs/2601.21337}, 
}

@inproceedings{whisper,
    author = {Radford, Alec and Kim, Jong Wook and Xu, Tao and Brockman, Greg and McLeavey, Christine and Sutskever, Ilya},
    title = {{Robust speech recognition via large-scale weak supervision}},
    year = {2023},
    publisher = {JMLR.org},
    booktitle = {Proceedings of the 40th International Conference on Machine Learning},
    articleno = {1182},
    numpages = {27},
    location = {Honolulu, Hawaii, USA},
    series = {ICML'23}
}

@misc{kyutai2025streaming,
    title={{Streaming Sequence-to-Sequence Learning with Delayed Streams Modeling}}, 
    author={Neil Zeghidour and Eugene Kharitonov and Manu Orsini and Václav Volhejn and Gabriel de Marmiesse and Edouard Grave and Patrick Pérez and Laurent Mazaré and Alexandre Défossez},
    year={2025},
    eprint={2509.08753},
    archivePrefix={arXiv},
    primaryClass={cs.CL},
    url={https://arxiv.org/abs/2509.08753}, 
}

@misc{sekoyan2025canary1bv2parakeettdt06bv3efficient,
    title={{{Canary-1B-v2} \& {Parakeet-TDT-0.6B-v3}: Efficient and High-Performance Models for Multilingual {ASR} and {AST}}}, 
    author={Monica Sekoyan and Nithin Rao Koluguri and Nune Tadevosyan and Piotr Zelasko and Travis Bartley and Nikolay Karpov and Jagadeesh Balam and Boris Ginsburg},
    year={2025},
    eprint={2509.14128},
    archivePrefix={arXiv},
    primaryClass={cs.CL},
    url={https://arxiv.org/abs/2509.14128}, 
}

@misc{microsoft2025phi4minitechnicalreportcompact,
    title={{Phi-4-Mini Technical Report: Compact yet Powerful Multimodal Language Models via Mixture-of-{LoRAs}}}, 
    author={Abdelrahman Abouelenin and Atabak Ashfaq and Adam Atkinson and Hany Awadalla and Nguyen Bach and Jianmin Bao and Alon Benhaim and Martin Cai and Vishrav Chaudhary and Congcong Chen and Dong Chen and Dongdong Chen and Junkun Chen and Weizhu Chen and Yen-Chun Chen and Yi-ling Chen and Qi Dai and Xiyang Dai and Ruchao Fan and Mei Gao and others},
    year={2025},
    eprint={2503.01743},
    archivePrefix={arXiv},
    primaryClass={cs.CL},
    url={https://arxiv.org/abs/2503.01743}, 
}

@misc{saon2025granitespeechopensourcespeechawarellms,
    title={{Granite-speech: open-source speech-aware {LLMs} with strong English {ASR} capabilities}}, 
    author={George Saon and Avihu Dekel and Alexander Brooks and Tohru Nagano and Abraham Daniels and Aharon Satt and Ashish Mittal and Brian Kingsbury and David Haws and Edmilson Morais and Gakuto Kurata and Hagai Aronowitz and Ibrahim Ibrahim and Jeff Kuo and Kate Soule and Luis Lastras and Masayuki Suzuki and Ron Hoory and Samuel Thomas and others},
    year={2025},
    eprint={2505.08699},
    archivePrefix={arXiv},
    primaryClass={eess.AS},
    url={https://arxiv.org/abs/2505.08699}, 
}

@misc{Silero_VAD,
    author = {{Silero Team}},
    title = {{Silero {VAD}: pre-trained enterprise-grade Voice Activity Detector ({VAD}), Number Detector and Language Classifier}},
    year = {2024},
    publisher = {GitHub},
    journal = {GitHub repository},
    howpublished = {\url{https://github.com/snakers4/silero-vad/tree/v5.1.2}},
    commit = {6478567951ae5c9979ad7b234185b5515f4be7a1}, 
    email = {hello@silero.ai}
}

@misc{canary_qwen,
  author = {{NVIDIA NeMo Team}},
  title = {{{Canary-Qwen-2.5B}: A Speech-Augmented Language Model}},
  year = {2025},
  publisher = {Hugging Face},
  journal = {Hugging Face Repository},
  howpublished = {\url{https://huggingface.co/nvidia/canary-qwen-2.5b}},
  note = {{Accessed: 2025-10-19}}
}

@misc{openai2025gptoss120bgptoss20bmodel,
      title={{gpt-oss-120b \& gpt-oss-20b Model Card}}, 
      author={Sandhini Agarwal and Lama Ahmad and Jason Ai and Sam Altman and Andy Applebaum and Edwin Arbus and Rahul K. Arora and Yu Bai and Bowen Baker and Haiming Bao and Boaz Barak and Ally Bennett and Tyler Bertao and Nivedita Brett and Eugene Brevdo and Greg Brockman and Sebastien Bubeck and Che Chang and Kai Chen and Mark Chen and others},
      year={2025},
      eprint={2508.10925},
      archivePrefix={arXiv},
      primaryClass={cs.CL},
      url={https://arxiv.org/abs/2508.10925}, 
}

@misc{singh2025openaigpt5card,
      title={{OpenAI GPT-5 System Card}}, 
      author={Aaditya Singh and Adam Fry and Adam Perelman and Adam Tart and Adi Ganesh and Ahmed El-Kishky and Aidan McLaughlin and Aiden Low and AJ Ostrow and Akhila Ananthram and Akshay Nathan and Alan Luo and Alec Helyar and Aleksander Madry and Aleksandr Efremov and Aleksandra Spyra and Alex Baker-Whitcomb and Alex Beutel and Alex Karpenko and Alex Makelov and others},
      year={2025},
      eprint={2601.03267},
      archivePrefix={arXiv},
      primaryClass={cs.CL},
      url={https://arxiv.org/abs/2601.03267}, 
}

@InProceedings{conneau2022fleurs,
    author={Conneau, Alexis and Ma, Min and Khanuja, Simran and Zhang, Yu and Axelrod, Vera and Dalmia, Siddharth and Riesa, Jason and Rivera, Clara and Bapna, Ankur},
    booktitle={2022 IEEE Spoken Language Technology Workshop (SLT)}, 
    title={{{FLEURS}: {FEW}-Shot Learning Evaluation of Universal Representations of Speech}}, 
    year={2023},
    volume={},
    number={},
    pages={798-805},
    doi={10.1109/SLT54892.2023.10023141}
}

@InProceedings{ardila2020common,
    author    = {Ardila, Rosana  and  Branson, Megan  and  Davis, Kelly  and  Kohler, Michael  and  Meyer, Josh  and  Henretty, Michael  and  Morais, Reuben  and  Saunders, Lindsay  and  Tyers, Francis  and  Weber, Gregor},
    title     = {{Common Voice: A Massively-Multilingual Speech Corpus}},
    booktitle      = {Proceedings of The 12th Language Resources and Evaluation Conference},
    month          = {May},
    year           = {2020},
    address        = {Marseille, France},
    publisher      = {European Language Resources Association},
    pages     = {4218--4222},
    url       = {https://www.aclweb.org/anthology/2020.lrec-1.520}
}

@inproceedings{panayotov2015librispeech,
    author={Panayotov, Vassil and Chen, Guoguo and Povey, Daniel and Khudanpur, Sanjeev},
    booktitle={2015 IEEE International Conference on Acoustics, Speech and Signal Processing (ICASSP)}, 
    title={{Librispeech: An {ASR} corpus based on public domain audio books}}, 
    year={2015},
    volume={},
    number={},
    pages={5206-5210},
    doi={10.1109/ICASSP.2015.7178964}
}

@misc{earnings22,
      title={{Earnings-22: A Practical Benchmark for Accents in the Wild}}, 
      author={Miguel Del Rio and Peter Ha and Quinten McNamara and Corey Miller and Shipra Chandra},
      year={2022},
      eprint={2203.15591},
      archivePrefix={arXiv},
      primaryClass={cs.CL},
      url={https://arxiv.org/abs/2203.15591}, 
}

@inproceedings{barker18,
  author={Jon Barker and Shinji Watanabe and Emmanuel Vincent and Jan Trmal},
  title={{The Fifth 'CHiME' Speech Separation and Recognition Challenge: Dataset, Task and Baselines}},
  year=2018,
  booktitle={Proc. Interspeech 2018},
  pages={1561--1565},
  doi={10.21437/Interspeech.2018-1768}
}

@InProceedings{carletta2006,
author="Carletta, Jean and Ashby, Simone and Bourban, Sebastien and Flynn, Mike and Guillemot, Mael and Hain, Thomas and Kadlec, Jaroslav and Karaiskos, Vasilis and Kraaij, Wessel and Kronenthal, Melissa and Lathoud, Guillaume and Lincoln, Mike and Lisowska, Agnes and McCowan, Iain and Post, Wilfried and Reidsma, Dennis and Wellner, Pierre",
editor="Renals, Steve and Bengio, Samy",
title="{{The AMI Meeting Corpus: A Pre-announcement}}",
booktitle="Machine Learning for Multimodal Interaction",
year="2006",
publisher="Springer Berlin Heidelberg",
address="Berlin, Heidelberg",
pages="28--39",
isbn="978-3-540-32550-5"
}

@misc{ldc2002t43,
  author       = {Linguistic Data Consortium},
  title        = {{2000 HUB5 English Evaluation Transcripts}},
  year         = {2002},
  publisher    = {Linguistic Data Consortium},
  address      = {Philadelphia},
  note         = {LDC2002T43, Web Download}
}

@misc{srivastav2025openasrleaderboardreproducible,
      title={{Open ASR Leaderboard: Towards Reproducible and Transparent Multilingual Speech Recognition Evaluation}}, 
      author={Vaibhav Srivastav and Steven Zheng and Eric Bezzam and Eustache Le Bihan and Adel Moumen and Sanchit Gandhi},
      year={2025},
      eprint={2510.06961},
      archivePrefix={arXiv},
      primaryClass={cs.CL},
      url={https://arxiv.org/abs/2510.06961}, 
}

\end{document}